\documentclass[runningheads]{llncs}
\usepackage{graphicx}

\usepackage{tikz}
\usepackage{comment}
\usepackage{amsmath,amssymb} 
\usepackage{color}

\usepackage[accsupp]{axessibility}  
\usepackage{wrapfig}
\usepackage{graphicx}
\usepackage{booktabs}
\usepackage{xcolor}
\usepackage{tabularx}
\usepackage{makecell}
\usepackage{cuted}  
\usepackage{capt-of}  
\usepackage{bbm}   
\usepackage{bbding}    
\usepackage{pifont}    
\usepackage{multirow}  
\usepackage{caption, subcaption, overpic, textpos}

\usepackage{hyperref} 
\hypersetup{ 
	colorlinks=true, 
	linkcolor=blue, 
	filecolor=blue, 
	citecolor = black,       
	urlcolor=blue, 
}

\begin{document}
\pagestyle{headings}
\mainmatter
\def\ECCVSubNumber{3659}  

\title{Towards Self-supervised and Weight-preserving Neural Architecture Search} 

%
\author{Zhuowei Li$^*$\inst{1} \and
Yibo Gao$^*$$^\dagger$\inst{2} \and Zhenzhou Zha$^\dagger$ \inst{3} \and Zhiqiang Hu \inst{4} \and Qing Xia \inst{4} \and Shaoting Zhang \inst{4} Dimitris N. Metaxas \inst{1}}
%
%
\institute{Rutgers Univeristy, USA, \email{zl502@cs.rutgers.edu} \and University of Electronic Science and Technology of China, CN \and Zhejiang University, CN \and SenseTime Research, CN}
\maketitle

\def\thefootnote{$*$}\footnotetext{Equal contributions.}
\def\thefootnote{$\dagger$}\footnotetext{This work was done during the internship at SenseTime.}

\begin{abstract}
Neural architecture search (NAS) algorithms save tremendous labor from human experts. Recent advancements further reduce the computational overhead to an affordable level. However, it is still cumbersome to deploy the NAS techniques in real-world applications due to the fussy procedures and the supervised learning paradigm. In this work, we propose the self-supervised and weight-preserving neural architecture search (SSWP-NAS) as an extension of the current NAS framework by allowing the self-supervision and retaining the concomitant weights discovered during the search stage. As such, we simplify the workflow of NAS to a one-stage and proxy-free procedure. Experiments show that the architectures searched by the proposed framework achieve state-of-the-art accuracy on CIFAR-10, CIFAR-100, and ImageNet datasets without using manual labels. Moreover, we show that employing the concomitant weights as initialization consistently outperforms the random initialization and the two-stage weight pre-training method by a clear margin under semi-supervised learning scenarios. Codes are publicly available at \url{https://github.com/LzVv123456/SSWP-NAS}. 

\end{abstract}

\section{Introduction}
\label{sec:intro}
The development of NAS algorithms save considerable time and efforts of human experts through automating the neural architecture design process. It has achieved state-of-the-art performances in a series of vision tasks including image recognition~\cite{ZophL16,nasnet,Amoba}, semantic segmentation~\cite{AutoDeepLab,nas_segmentation} and object detection~\cite{DBLP:journals/corr/abs-1906-04423,Ghiasi2019NASFPNLS}. Recent advances on weight-sharing NAS~\cite{EfficientNAS} and differentiable NAS~\cite{DARTS,snas} further reduce the searching cost from thousands of GPU-days to a couple. 

\begin{figure}[t!]
	\centering
	\includegraphics[width=0.95\textwidth]{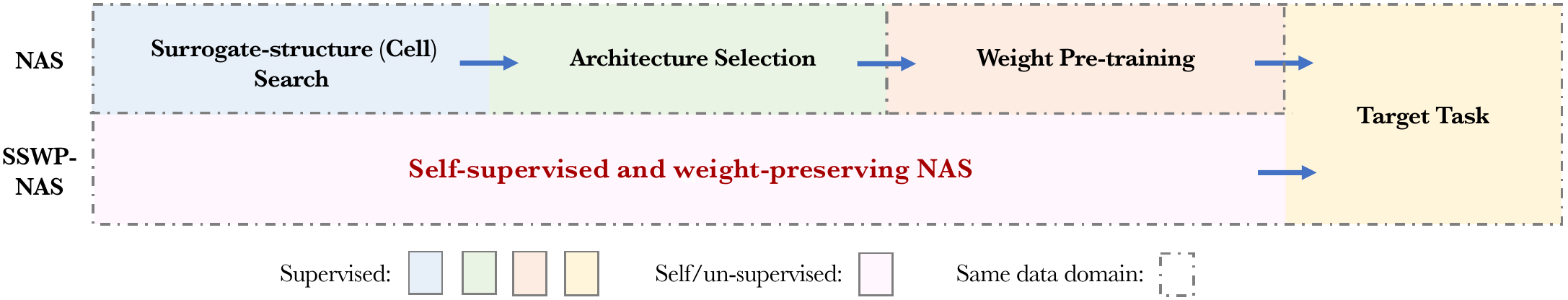}
	\caption{Overview for the regime of general NAS and the proposed SSWP-NAS. \label{fig:overview}}
\end{figure}

Despite the significant computational reduction made by current NAS methods, it is still cumbersome to deploy the NAS techniques in real-world applications due to the fussy procedures. As shown in Fig.~\ref{fig:overview}, a typical workflow of NAS consists of the surrogate-structure search and the architecture selection two steps to acquire the architecture. Then a standalone procedure of weight pre-training need to be taken before transferring the architecture to downstream tasks. It is non-trivial to pre-train a network, taking even more time than the searching process. Besides, existing NAS workflows largely rely on manual annotations, making the domain-specific NAS even more unwieldy. 

Driven by the inconvenience of the current NAS paradigm, we propose a new framework, namely self-supervised and weight-preserving neural architecture search (SSWP-NAS), as an extension of the current NAS methodology with the following two prominent properties: (1) SSWP-NAS is self-supervised so that it does not rely on manual signals to perform optimization. This property also removes the dependency on proxy-datasets (e.g. ImageNet~\cite{ImageNet}). (2) SSWP-NAS has the weight-preserving property, which means the concomitant weights generated during the search process can be retained and serve as initialization to benefit transfer learning. This property simplifies the current NAS workflow from the two-stage fashion to the one-stage. To achieve weight-preserving, we align the dimensionality of the network used during the search and train stage and leverage stochastic operation sampling strategy to reduce the memory footprint. To remove the dependency on manual labels, we probe how the designed searching process copes with the self-supervised learning objective. We also observe a persistent optimization challenge dubbed \textit{network inflation issue} caused by the inconsistent optimization targets and the stochastic strategy. To overcome this challenge, we propose the forward progressive prune (FPP) operation that gradually bridges the gap between the optimization targets and reduces the extend of stochastic operations. 

In experiments, the searched architecture using SSWP-NAS achieve state-of-the-art accuracy on CIFAR-10 ($2.41\%$ error rate), CIFAR-100 ($16.47\%$ error rate), and ImageNet ($24.3\%$ top-1 error rate under restricted resources) without using manual labels. Besides, concomitant weights as initialization consistently outperforms the random initialization and the two-stage weight pre-training method by a clear margin under semi-supervised learning scenarios. Moreover, we show that self-supervised learning objective consistently outperforms the supervised counterpart in our framework and FPP is beneficial under both supervised and self/un-supervised learning objectives. Comprehensive ablation studies have also been conducted towards the proposed designs. Our main contributions can be summarized in three-fold:
\begin{itemize}
	\item We propose the SSWP-NAS that enjoys the self-supervised learning and weight-preserving property. It simplifies the current NAS workflow from the two-stage manner to the one stage. As a by-product, SSWP-NAS is also proxy-free, which means it relies on neither the surrogate structure nor the proxy-dataset.
	\item
	We propose the FPP to address the network inflation issue that occurs during the designed weight-preserving search process. We empirically show that FPP is beneficial under both supervised and self/un-supervised signals. 
	\item  SSWP-NAS searches the architecture and generate the pre-train weights concurrently while achieving state-of-the-art performance regarding the quality of both the architecture and the pre-train weights.
\end{itemize}

\section{Related Work}
\textbf{Neural Architecture Search.} Early works for NAS mainly leverage on reinforcement learning (RL)~\cite{ZophL16,nasnet} or evolutionary algorithms (EA)~\cite{Amoba,HierarchicalNAS} to optimize a controller that samples a sequence of discrete operations to form the architecture. This straightforward implementation consumes tremendous computational resources. As a remedy, the following works rely on the weight-sharing~\cite{EfficientNAS,DARTS} methods and surrogate structures~\cite{nasnet,DARTS} to reduce the computational overhead. DARTS~\cite{DARTS} further simplify the NAS framework by relaxing the search space from discrete to continuous. This \textit{One-shot} searching framework then becomes popular in the NAS domain due to its simplicity and efficiency. The proposed SSWP-NAS also inheres to this line of research. While the original DARTS design is observed to suffer from performance degeneration~\cite{PDARTS} and mode collapse~\cite{darts_plus} issue. To this end, DARTS+~\cite{darts_plus} introduces the early stop to suppress the over-characterized non-parameterized operations. P-DARTS~\cite{PDARTS} tries to alleviate the performance drop by gradually increasing the depth of the surrogate structure. ProxylessNAS~\cite{ProxylessNAS} first achieves the differentiable architecture search without a surrogate structure. Despite the advancements made above methods, they still rely on supervised signals to perform optimization, and none of them are able to preserve the concomitant weights.

\noindent\textbf{Self-supervised Learning.} Not until recently, self-supervised learning largely relies on heuristic pretext tasks~\cite{Context-Prediction,Jigsaw,Split-Brain,Local-Aggregation} to form the supervision signal, and their performance lags behind a lot compared with the supervised counterpart. Emerging of the contrastive-based self-supervised learning largely close the gap between the self-supervised and supervised weight pre-training~\cite{CPC,Mutiview-CPC,MOCO,SimCLR,PCL}. It encourages a pulling force within a positive pair and pushing away negative pairs through minimizing a discriminative loss function. BYOL~\cite{BYOL} and SimSiam~\cite{simsiam} also demonstrate that the negative pairs are not necessary. In this work, we investigate how state-of-the-art self-supervised learning methods cope with the weight-preserving network search.

\noindent\textbf{Self/un-supervised NAS.} Most recently, some other works also explore the self/un-supervised learning objective under the NAS framework. UnNAS~\cite{UNNAS} explores how different pretexts tasks can replace the supervised discrimination task. It shows that labels are not necessary for NAS, and metrics used in pretext tasks can be a good proxy for the structure selection. RLNAS~\cite{RLNAS} shifts from the performance-based evaluation metric to the convergence-based metric, and it uses the random labels to generate supervision signals. Among self/un-supervised NAS works, SSNAS~\cite{SSNAS} and CSNAS~\cite{CSNAS} are most similar to our SSWP-NAS as they also explore the contrastive learning under the NAS framework. Nevertheless, they only consider the picture from an architecture optimization perspective, and their frameworks still fall into proxy-based searching. Dissimilarly, we are searching for the architecture and concomitant weights as integrity.

\section{Methodology}
In this section, we first introduce the prior knowledge about differential NAS which serves as the foundation of our framework. Then we detail how to extend the differential NAS towards the weight-preserving search and self-supervised optimization. Afterward, we investigate the network inflation issue and propose our solution. Finally, we demonstrate how to search a network (architecture plus weights) using SSWP-NAS.

\subsection{Preliminary: differentiable NAS}
\label{priors}
Neural architecture search (NAS) task is generally formulated as a bi-level optimization task~\cite{Hierarchical-optimization,bilevel-optimization} where the upper-level variable $\alpha$ refers to the architecture parameters and lower-lever variable $w$ represents the operation parameters:
\begin{equation}
	\min_{\alpha} \mathcal{L}_{val}(w^*(\alpha), \alpha)
\end{equation}
\begin{equation}
	s.t. \;\;\; w^*(\alpha) = argmin_w \; \mathcal{L}_{train}(w, \alpha)
\end{equation} 
In practice, two sets of parameters are optimized in an alternative manner that temporarily reduces the bi-level optimization to a single-level optimization. Differential NAS (DARTS)~\cite{DARTS} further includes the architecture parameters directly to the computational graph through relaxing the search space from discrete to continuous. Thus it can effectively evolve both architecture structure and operation weights leveraging the stochastic gradient descent~\cite{sgd} techniques. 

Inspired by the success of manually-designed structural motifs, NAS methods also shift the searching target from the intact architecture to a cell structure~\cite{nasnet,pnas,Amoba} which is then used to stack the final architecture repetitively. DARTS~\cite{DARTS} also adopts this strategy by first searching a cell unit using a light-weight proxy architecture followed by constructing the final architecture with the searched cell structure. And the DARTS search space~\cite{DARTS} represented by a cell can be interpreted as a directed acyclic graph (DAG) which consists of $7$ nodes and $14$ edges. For each edge, it is associated with a collection of candidate operations $\mathcal{O}$ weighted by a real-valued vector $\alpha^{(i,j)}$ of size $ M=|\mathcal{O}|$. And the information flow $f_{i,j}$ between node $x_i$ and $x_j$ is defined as:
\begin{equation}
	f_{i,j}(x_i) = \sum_{o\in\mathcal{O}}\frac{exp(\alpha_o^{(i,j)})}{\sum_{o'\in\mathcal{O}}exp(\alpha_{o'}^{(i,j)})} o(x_i)
\end{equation}
where $i<j$ and an intermediate node is computed based on all its predecessors: $x_j = \sum_{i<j}f_{i,j}(x_i)$. The final output of a cell is the concatenation of $4$ intermediate nodes (except $2$ input nodes and $1$ output node) over the channel dimension. Here we refer the DARTS~\cite{DARTS} paper for more details. 

\subsection{Towards Weight-preserving}
One of the most prominent challenges in NAS is the surge of the memory footprint and the computational overhead. As a remedy, the decomposition from the whole architecture to the cell structure and the surrogate architecture searching is proposed to function as a trade-off between accuracy and efficiency. While reducing the computational overhead, the prevalence of such proxy strategy excludes the possibility of the weight-preserving property from the very beginning. Due to the non-identical structures used during the search and the train stage, concomitant weights discovered in the search phase are abandoned, and only the searched cell structure is reserved for the downstream application. 

To retain the concomitant weights, we first build our target architecture using the DARTS~\cite{DARTS} search space, which is well-established and contains abundant sub-graphs~\cite{bench101}. However, different from the DARTS and following advances~\cite{PDARTS,darts_plus}, we align the dimensionality (widths and depths) of the architecture utilized during the search and the train stage. We further allow different cell structures at each level of the architecture. That is, instead of searching a single cell structure as the building unit, we search a set of cells to form the final architecture directly. As clued in section~\ref{priors}, this over-parameterized formation will consume $ M=|\mathcal{O}|$ times memory footprint compared with a compact architecture. Disregarding the proxy strategy that contradicts the weight-preserving property in nature, we resort to stochastic algorithms~\cite{snas,ProxylessNAS,PC-DARTS} to cut down memory usage. Here, we adopt the path-binarization strategy, which is first proposed in ProxylessNAS~\cite{ProxylessNAS} to overcome the accuracy degeneration issue caused by the depth-gap between the surrogate and the final architecture. Specifically, only a single operation on an edge is sampled and activated stochastically according to a learned distribution at each iteration. As such, the memory footprint during the search is reduced to the same magnitude as a compact architecture. 

Through renouncing the proxy strategy and adopting the stochastic technique, we meet the indispensable conditions towards the weight-preserving. While in practice, we observe a consistently undesirable edge gained by the skip-connection operation. Similar phenomenons have also been observed in previous supervised and proxy-based search~\cite{darts_plus,PDARTS,Shu2020Understanding}. Such over-ratings for non-parameterized operations not only hinder the general quality of the structure, but also result in insufficient updates of parameterized operations during the lower-level optimization. To this end, we introduce a non-parameterized operation dropout ($p=0.2$ across our experiments) during the lower-level optimization. By doing so, we implicitly regularize the importance of non-parameterized operation and increase the sampling odds for parameterized operation without direct interference with the learned upper-level parameter distribution.

\begin{figure}[t]
	\centering
	\includegraphics[width=0.95\textwidth]{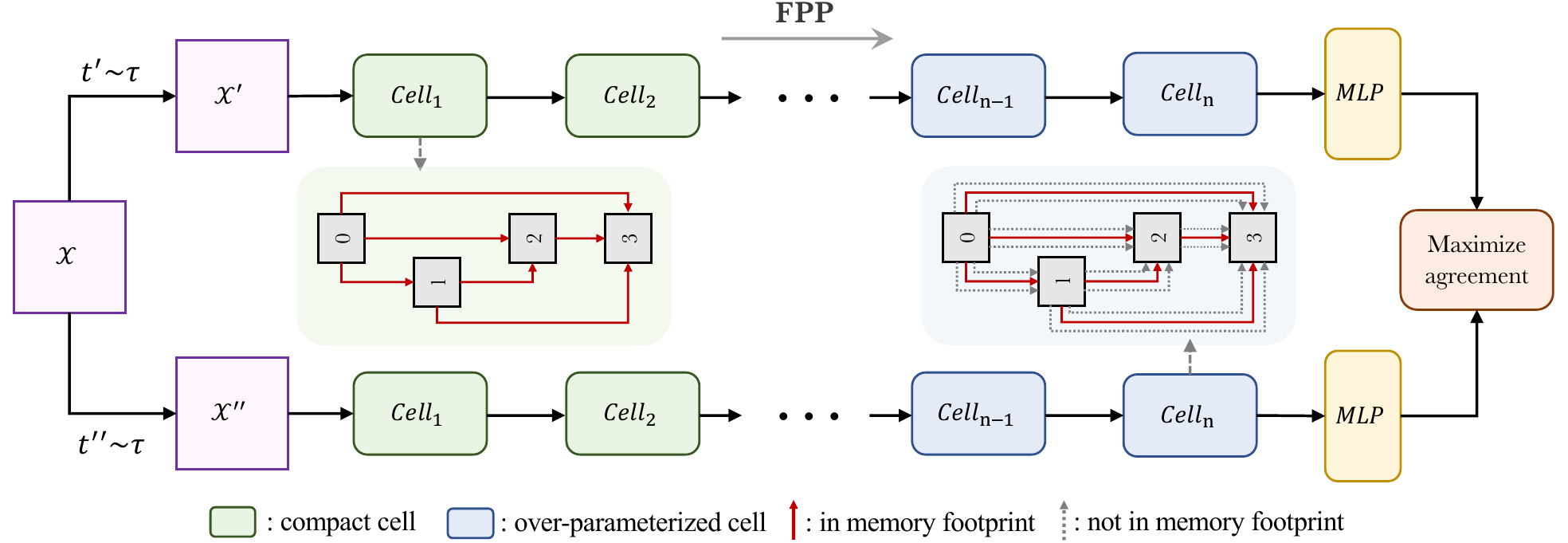}
	\caption{The proposed SSWP-NAS framework. $\tau$ is a collection of data augmentations, $t'$ and $t''$ are two transformations sampled from $\tau$ that transfer image $x$ to two different views $x'$ and $x''$, respectively. Cell-structure plotted here is simplified for the interpretation purpose.}
	\label{fig:framework}
\end{figure}

\subsection{Towards Self-supervised Learning}
It is conceptually straightforward to remove the dependency of the manual annotation by replacing the supervised signal with a self/un-supervised counterpart. Following the common practice of NAS, we are seeking a favorable architecture through modeling the conditional probability (discriminative model) rather than learning the joint distribution (generative model). As such, we subscribe our exploration for the learning objective to the discriminative sub-field. And among the bag of self/un-supervised discriminative pretext tasks, contrastive learning based methodologies~\cite{CPC,Mutiview-CPC,MOCO,SimCLR,BYOL} demonstrate an outstanding representation learning ability over the peers. So we further focus our attention on probing how the contrastive learning copes with the designed weight-preserving search process. 

Given that the concept of the positive and negative pair lies at core of the contrastive learning, we study two representative methods, SimCLR~\cite{SimCLR} and BYOL~\cite{BYOL} which are formulated with and without negative pairs, respectively. According to our pilot trials, BYOL only achieves $3.05\%$ top-1 error rate on cifar-10 while SimCLR achieving $2.56\%$. BYOL lags behind a lot. We observe that the failure of the BYOL framework originates from the incompatible patterns between the stochastic operation sampling and exponential moving average (EMA). BYOL essentially forms a teacher-student pair using the current snapshot of the operation weights $\theta_t$ and its corresponding EMA counterpart $\theta_t'=\tau\theta'_{t-1}+(1-\tau)\theta_t$ where $\tau$ is a smoothing factor. Then BYOL employees a symmetric consistency loss between the pair as the supervision signal. However, stochastic operation sampling turns the parameter distribution $\theta_t$ into $\xi_t$ where $\xi$ and $\theta$ are different operation distributions. As a result, this inconsistency causes the mode collapse of the EMA, and thus fails the BYOL framework. As such, we take the strategy from SimCLR~\cite{SimCLR} where positive pairs are formed by two transformations of the same image and negative pairs are constructed using different images sampled from a mini-batch. The intact search architecture is then optimized in an end-to-end manner through minimizing the infoNCE loss~\cite{infoNCE,infoNCE2} with mini-batch size $N$:
\begin{equation}
	\mathcal{L}_{batch} = \sum^{2N}_{i=1}\mathcal{L}_{i}
\end{equation}
\begin{equation}
	\mathcal{L}_i = -\log{\frac{\exp{(sim(\boldsymbol{z}_i,\boldsymbol{z}_{j(i))}/\tau)}}{\sum^{2N}_{k=1}\mathbbm{1}_{[k \neq i]}\exp(sim(\boldsymbol{z}_i,\boldsymbol{z}_k))/\tau)}}
\end{equation}
Here, $\boldsymbol{z}_i = P(E(\tilde{\boldsymbol{x}}_{i}))$. $E(\cdot)$ denotes the searching encoder architecture and $P(\cdot)$ is a projection neck (multi-layer perception) added at end of the searching structure. The Subscript $i$ and $j(i)$ refers to two different views of the same image. And $sim(\cdot)$ measures the cosine similarity between two given vectors. Fig.~\ref{fig:framework} displays the construction of our framework. This self-supervision pipeline is totally orthogonal to the search space and search process design. It is worth noting that $E_t(\cdot)\neq E_{t-1}(\cdot)$ at different time steps and $P(\cdot)$ is only employed during the search stage and being replaced by a linear classification layer for supervised train stage. By leveraging this self-guided signal generation method, SSWP-NAS gets rid of the dependency on manual labels.
\subsection{Network Inflation Challenge}
Despite a neat couple made by the weight-preserving search and the self-supervised optimization, it is challenging to optimize such a framework in practice. We observe that the difficulty comes from the combination of the over-parameterized structure and the stochastic operation sampling strategy, which in together we referred to as \textit{network inflation} issue.

From the macro perspective, we are optimizing two sets of over-parameterized variable distributions $\alpha$ and $w$ alternatively through the whole searching phase. At the end of search process, differentiable NAS~\cite{DARTS} relies on the non-linear prune operation to approximate the compact variable $\tilde{\alpha}^*$ and $\tilde{w}^*$ using the over-parameterized variable  $\alpha^*$ and $w^*$. This leaping relaxation essentially relies on the good generalization ability of the hierarchical convolutional structures. From a micro view, a hierarchical convolutional structure can be viewed as a sequential model. Given a random tie-breaking index $i$ in a $N$-layer structures at time step $t$, we are maximizing posterior probability $\arg\max_{w_{i\sim N}}p(w_{i\sim N}|w^t_{1\sim i-1}, D)$ where $D$ is the data distribution and subscript $a\sim b$ denotes layer index from $a$ to $b$. Due to the stochastic sampling strategy, conditions $w^{t-1}_{1\sim i-1}\neq w^{t}_{1\sim i-1}$ while $w_{i\sim N}$ can be deemed as unchanged. Even though stochastic algorithm theoretically guarantees the same global convergence, if it exists, this non-stationary condition increases the difficulty of optimization at each time step. Here we take the lower-level variable $w$ as an example, and the upper-level variable follows the similar formulation.

To handle the network inflation challenge, we propose a simple and effective solution, namely \textit{forward progressive prune (FPP)}. Different from the general differentiable NAS pipeline where an architecture prune process is conducted for the target cell structure at the end of the search stage. We impose a cell-level progressive prune in the forward propagation direction during the search stage. At each prune step, we prune all edges contained in a cell, and for each edge, we only keep the operation with highest learned credit. By doing so, we reformulate the optimization target as  $\arg\max_{w_{i\sim N}}p(w_{i\sim N}|\tilde{w}^*_{1\sim i-1}, D)$ where the condition $\tilde{w}^*_{1\sim i-1}$ is fixed and we transfer the optimization target from $p(w_{1\sim N})$ to $p(w_{i\sim N}|\tilde{w}^*_{1\sim i-1})$ which is closer to the final goal $p(\tilde{w}_{1\sim N}^*)$. As such, FPP gradually aligns the searching optimization target with the final objective. And it fixes the non-stationary conditions at some point during the search and allows the following layers to adjust according to the preceding layers. The forward propagation direction design follows the common acknowledgment that shallow layers of a CNN architecture capture the low-level features which are easier to learn and the deeper layers grab more semantics which rely on low-level features to compose.

\subsection{Searching with SSWP-NAS}
\begin{figure}[t]
	\centering
	\includegraphics[width=0.7\textwidth]{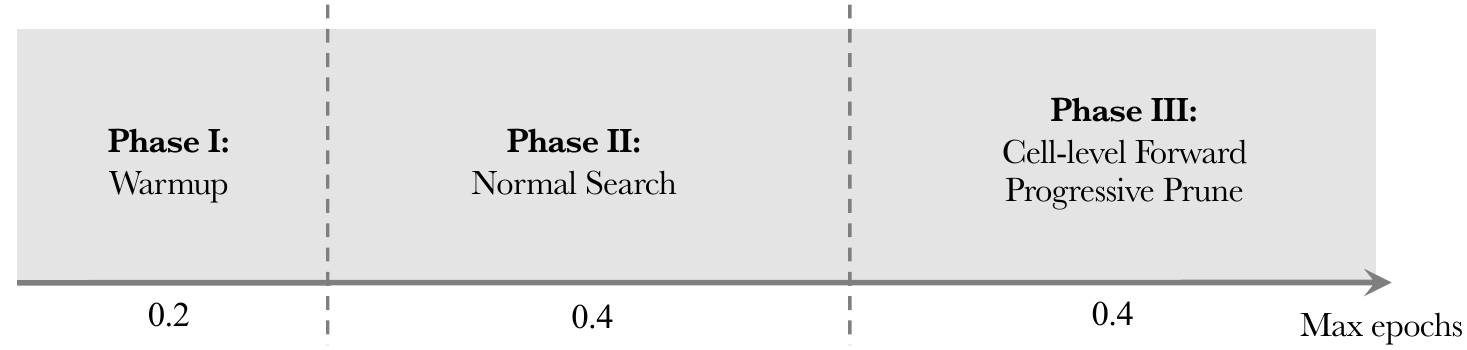}
	\caption{Different phases during the search stage.}
	\label{fig:search_process}
\end{figure}

SSWP-NAS is a proxy-free search, so one can search the target architecture, not surrogate architecture, directly on target data domain without using labels. As depicted in Fig.~\ref{fig:search_process}, the search stage of SSWP-NAS is divided into three phases. At the warm-up phase, we only update operation parameters to allow a better initialization of parameterized operations. Then we update both operation parameters and architecture parameters alternatively as the standard differentiable NAS. Finally, we start the FPP phase, in which we progressively perform cell-level prune. The only extra hyper-parameter we introduce is the time span of different phases. We empirically suggest that the split $[0.2, 0.4, 0.4]$ works well in general. Given the proportion for phase III, the time step for pruning of two adjacent cells is calculated as $T_{step}=\frac{max \; epoch \times FPP \; ratio}{cell \; num}$.

\section{Experiments}
Experimental section is organized as following. We first provide the general settings used across our experiments. In the second part, we isolate the concomitant weights and benchmark the architecture quality searched using SSWP-NAS. Then we conduct comprehensive ablation studies towards our designs. Finally, we thoroughly study the effectiveness of concomitant weights. Detailed searched architecture are attached in \textit{Appendix C}.

\subsection{Experimental Settings} 
Following the well-established benchmark, we conduct our experiments on CIFAR-10/100~\cite{CIFAR} and ImageNet~\cite{ImageNet} three datasets. We search an architecture consisting $20$ cells with $36$ initial channels for $300$ epochs using mini-batch size $96$ on CIFAR-10/100. For ImageNet, we search a structure that contains $14$ cells with $48$ initial channels for $100$ epochs using mini-batch size $256$. The magnitudes of searched architectures are kept the same as the final architecture trained in DARTS~\cite{DARTS}. All hyper-parameters used during the search and train stage, except our proposed designs, are inherited from DARTS~\cite{DARTS} without extra tricks. We attach detailed hyper-parameters in \textit{Appendix A}. And $1_{st}$ order optimization is employed across all settings. Experiments related to CIFAR-10/100~\cite{CIFAR} are conducted on a single Nvidia A$100$ $40$GB and it scales up to $4$ for ImageNet~\cite{ImageNet}. 

\begin{table*}[t!]
	\caption{Comparisons between SSWP-NAS and state-of-the-art methods on CIFAR-10/100. Here the search cost only counts the time used for the search. proxy-based methods commonly need another architecture selection procedure which generally costs another $1$ GPU day~\cite{DARTS}. While this procedure is not needed in our framework.}
	\label{tab:cifar}
	\begin{center}
		\resizebox{0.95\textwidth}{!}{%
			\begin{tabular}{lcccccc}
				\hline\\[-2.5ex]
				\multirow{2}{*}{\textbf{Architecture}} &\multicolumn{2}{c}{\textbf{Test Error (\%)}} &\textbf{Params} &\textbf{Search Cost} &{\textbf{Search}} &{\textbf{Search}}\\\cline{2-3}\\[-2.5ex]
				&\textbf{CIFAR-10} &\textbf{CIFAR-100} &\textbf{(M)} &\textbf{(GPU days)} &\textbf{Type}  &\textbf{Method} \\
				\hline\\[-2.5ex]
				DenseNet-BC\cite{densenet}             &$3.46$   &$17.18$  &$25.6$   &-      &supervised         &manual      \\
				\hline\\[-2.5ex]
				NASNet-A\cite{nasnet}                  &$2.65$   &-      &$3.3$    &$1800$   &supervised         &RL          \\
				AmoebaNet-A\cite{Amoba}            &$3.34\pm0.06$   &-      &$3.2$    &$3150$   &supervised         &evolution   \\
				AmoebaNet-B\cite{Amoba}            &$2.55\pm 0.05$   &-      &$2.8$    &$3150$   &supervised         &evolution   \\
				PNAS\cite{pnas}                        &$3.41\pm 0.09$   &-      &$3.2$    &$225$    &supervised         &SMBO        \\
				Hireachical Evolution~\cite{HierarchicalNAS}   &$3.75\pm 0.12$  &-  &$15.7$  &$300$  &supervised  &evolution \\
				ENAS\cite{EfficientNAS}                        &$2.89$   &-      &$4.6$    &$0.5$    &supervised         &RL          \\
				NAONet\cite{naonet}                    &$3.18$   &$15.67$  &$10.6$   &$200$    &supervised         &NAO         \\
				\hline\\[-2.5ex]
				DARTS (1st order)\cite{DARTS}        &$3.0\pm 0.14$  &$17.76$  &$3.3$    &$1.5$  &supervised         &gradient    \\
				DARTS (2nd order)\cite{DARTS}       &$2.76\pm 0.09$  &$17.54$  &$3.3$    &$4.0$  &supervised         &gradient    \\
				SNAS (moderate)\cite{snas}             &$2.85\pm 0.02$   &-      &$2.8$    &$1.5$ &supervised         &gradient    \\
				ProxylessNAS-G\cite{ProxylessNAS}      &$2.08$   &-      &$5.7$    &$4.0$    &supervised         &gradient    \\
				P-DARTS\cite{PDARTS}                   &$2.50$   &$16.55$  &$3.4$    &$0.3$    &supervised         &gradient    \\
				PC-DARTS\cite{PC-DARTS}                 &$2.57\pm 0.07$   &-      &$3.6$    &$0.1$    &supervised         &gradient    \\
				BayesNAS\cite{Bayes_nas}                      &$2.81\pm 0.04$   &-  &$3.4$     &$0.2$   &supervised    &gradient    \\
				CSNAS$_{N=5}$\cite{CSNAS}                      &$2.66\pm 0.07$   &-      &$3.4$    &$1.0$   &self-supervised    &SMBO-TPE    \\
				SSNAS\cite{SSNAS}                      &$2.61$   &$16.64$  &-      &-   &self-supervised    &gradient    \\
				\hline\\[-2.5ex]
				\textbf{SSWP-NAS}$_{e=300}$$^{\dagger}$      &$2.56\pm 0.07$   &$17.27$  &$4.0$   &$1.0$  &self-supervised    &gradient    \\
				\textbf{SSWP-NAS}$_{e=500}$$^{\dagger}$       &$\mathbf{2.41\pm 0.07}$   &$\mathbf{16.47}$  &$3.8$    &$1.8$   &self-supervised    &gradient    \\
				\hline\\[-1ex]
				\multicolumn{7}{l}{\small $^{\dagger}$: run $5$ times with different seeds.}
			\end{tabular} %
		}
	\end{center}
\end{table*}

\subsection{Benchmarking SSWP-NAS}
\textbf{CIFAR-10/100.} We search SSWP-NAS for $300$ ($e=300$) and $500$ ($e=500$) epochs on CIFAR dataset. Then we train the searched architecture from scratch in a supervised manner. As shown in Table~\ref{tab:cifar}, SSWP-NAS$_{e=300}$ achieves state-of-the-art performance and outperforms DARTS ($1_{st}$ order) significantly by searching the same epochs. When we extend the searching duration from $300$ epochs to $500$ epochs, SSWP-NAS surpasses existing methods by a clear margin. 

\noindent \textbf{ImageNet.} We search $50$ and $100$ epochs on ImageNet~\cite{ImageNet} and then train the searched structure for $250$ epochs. As displayed in Table~\ref{tab:imagenet}, SSWP-NAS also achieves state-of-the-art accuracy on ImageNet under limited budgets. One potential drawback is that SSWP-NAS takes a relative longer time to search when considering the architecture solely as we are directly searching for the final structure instead of a surrogate structure. However, the weight-preserving property offsets this computational overhead as all other methods need an extra non-trivial pre-train step. By achieving state-of-the-art performance on both CIFAR and ImageNet datasets, it also manifests the generality of the proposed SSWP-NAS framework.

\begin{table*}[t!]
	\caption{Comparison with state-of-the-art architectures on ImageNet (restricted resources)}
	\label{tab:imagenet}
	\begin{center}
	\resizebox{0.95\textwidth}{!}{%
		\begin{tabular}{lccccccc}
			\hline\\[-2.5ex]
			\multirow{2}{*}{\textbf{Architecture}} &\multicolumn{2}{c}{\textbf{Test Error (\%)}} &\textbf{Params} & $\times$+ &\textbf{Search Cost} &{\textbf{Search}} &{\textbf{Search}}\\\cline{2-3}\\[-2.5ex]
			&\textbf{Top-1} &\textbf{Top-5} &\textbf{(M)} &\textbf{(M)} &\textbf{(GPU days)} &\textbf{Type} &\textbf{Method} \\
			\hline\\[-2.5ex]
			Inception-v2~\cite{Inception-v2}  &$25.2$  &$7.8$  &$11.2$  &-  &-  &-  &manual  \\
			MobileNet-v3 (Large $1.0$)~\cite{mobilenet-v3}  &$24.8$  &-  &$5.4$  &$219$  &-  &-  &manual  \\
			ShuffleNet(2$\times$)-v2~\cite{shufflenet-v2}  &$25.1$  &-  &$\approx 5$  &$591$  &-  &-  &manual  \\
			\hline\\[-2.5ex]
			NASNet-A~\cite{nasnet}  &$26.0$  &$8.4$  &$5.3$  &$564$  &$2000$  &supervised  &RL  \\
			AmoebaNet-A~\cite{Amoba}  &$25.5$  &$8.0$  &$5.1$  &$555$  &$3150$  &supervised  &evolution  \\
			AmoebaNet-B~\cite{Amoba}  &$26.0$  &$8.5$  &$5.3$  &$555$  &$3150$  &supervised  &evolution  \\
			PNAS~\cite{pnas}  &$25.8$  &$8.1$  &$5.1$  &$588$ &$255$  &supervised  &SMBO  \\
			DARTS (2nd order) ~\cite{DARTS} &$26.7$  &$8.7$  &$4.7$  &$574$ &$4$  &supervised  &gradient  \\
			P-DARTS~\cite{PDARTS}  &$24.1$  &$7.3$  &$5.4$  &$597$ &$2.0$  &supervised  &gradient  \\
			PC-DARTS~\cite{PC-DARTS}  &$24.2$  &$7.3$  &$5.3$  &$597$ &$3.8$  &supervised  &gradient  \\
			ProxylessNAS~\cite{ProxylessNAS}  &$24.9$  &$7.5$  &$7.1$  &$465$ &$8.3$  &supervised  &gradient  \\
			\hline\\[-2.5ex]
			SSNAS~\cite{SSNAS}  &$27.8$  &$9.6$  &$5.2$  &- &-  &self-supervised  &gradient  \\
			CSNAS$_{N=5}$~\cite{CSNAS}  &$25.8$  &$8.3$  &$5.1$  &$590$ &$2.5$  &self-supervised  &SMBO-TPE\\
			\hline\\[-2.2ex]
			\textbf{SSWP-NAS$_{e=50}$}  &$24.8$ &$7.8$ &$5.0$  &$597$ &$3.5$  &self-supervised  &gradient\\
			\textbf{SSWP-NAS$_{e=100}$}  &$24.3$ &$7.5$ &$4.9$  &$595$ &$7$  &self-supervised  &gradient\\
			\hline\\[-1ex]
		\end{tabular} %
	}
\end{center}
\end{table*}

\subsection{Ablation Study}
For simplicity, we abbreviate self-supervised learning and supervised learning as SSL and SL, respectively, in the following sections.

\noindent\textbf{SSL vs. SL.} In order to substantiate the effectiveness of SSL, we compare it with the SL (cross-entropy loss function). By keeping all other settings the same, we switch between SL and SSL objectives. As shown in Table~\ref{tab:ssl+fpp}, SSL outperforms SL significantly in our framework. This result not only exhibits the inessentiality of human annotations for architecture search, but also suggests that the manual interference may function adversely by limiting the optimization manifold. SSL objective may support learning a more generic structure for feature extraction. Beyond this frank improvement, we further show that the SSL searched architecture boosts the self-supervised weight pre-training in \textit{Appendix B}

\noindent\textbf{Effectiveness of FPP.} To demonstrate the effectiveness of the FPP module, we implement SSWP-NAS with and without FPP. When disregarding FPP, we remove phase III during the search and adjust the ratio of phases I and II to $0.2$ and $0.8$ accordingly. Then the architecture prune process is conducted once for all cells after the search stage. Besides the original FPP, we also implement a reversed version, namely backward progressive pruning (BPP), that starts pruning from the last cell and propagates backwardly. According to Table~\ref{tab:ssl+fpp}, FPP consistently improves the qualities of the searched architectures under both SSL and SL scenarios. BPP, on the contrary, even degenerates the performances. This result coincides with our knowledge that shallow layers extract low-level features, which are easy to learn. In contrast, deeper layers capture higher-level semantic features and leverage the low-level features to compose.
\begin{table*}[t!]
	\centering
	{
		\caption{Ablation studies on learning objective, FPP, skip-connection dropout and proxy-free search.}
		\label{tab:ablation}
		\begin{minipage}[t]{0.47\textwidth}{\begin{center}
			\subfloat[
			\textbf{SSL vs. SL and FPP.} Ablations for learning objectives and FPP.
			\label{tab:ssl+fpp}
			]{
			\resizebox{1.0\linewidth}{!}{%
			\begin{tabular}{l c c c}
				\hline\\[-2.5ex]
				Name  & Test Error (\%)  & Dataset & Search Epoch \\
				\hline\\[-2.5ex]
				SL  & $2.92$ & CIFAR-10 & $300$ \\
				SL+BPP  & $2.93$ & CIFAR-10 & $300$ \\
				SL+FPP  & $2.77$ & CIFAR-10 & $300$  \\
				SSL  & $2.65$ & CIFAR-10 & $300$  \\
				SSL+BPP  & $2.73$ & CIFAR-10 & $300$  \\
				\hline\\[-2.5ex]
				\textbf{SSL+FPP}  & $\mathbf{2.56}$ & CIFAR-10 & $300$ \\
				\hline
			\end{tabular}%
		}
	}
\\
	\subfloat[
	\textbf{FPP time span.} Abations for different ratios of searching phases.
	\label{tab:fpp_ratio} 
	]{
		\resizebox{1.0\linewidth}{!}{%
			\begin{tabular}{l c c c c}
				\hline\\[-2.5ex]
				Name  & Test Error (\%) & Ratio  &Dataset  &Seach Epoch\\
				\hline\\[-2.5ex]
				shorter & $2.65$ &$[0.2, 0.6, 0.2]$   &CIFAR-10  &$300$\\
				longer  & $2.68$ & $[0.2, 0.2, 0.6]$  &CIFAR-10  &$300$\\
				\textbf{default} & $\mathbf{2.56}$  & $[0.2, 0.4, 0.4]$  &CIFAR-10  &$300$\\
				\hline
			\end{tabular}%
		}
	}
		\end{center}}\end{minipage}
	\hspace{1em}
	\begin{minipage}[t]{0.47\textwidth}{\begin{center}
				
	\subfloat[
	\textbf{Dropout for skip-connection.} Abations on different dropout rate for skip connections.
	\label{tab:dropout} 
	]{
		\resizebox{1.0\linewidth}{!}{%
			\begin{tabular}{l c c c c}
				\hline\\[-2.5ex]
				Name & Test Error (\%) &Dataset  &Seach Epoch\\
				\hline\\[-2.5ex]
				no drop  & $2.63$  &CIFAR-10  &$300$\\
				0.5 drop & $2.68$   &CIFAR-10  &$300$\\
				\textbf{0.2 drop} & $\mathbf{2.56}$   &CIFAR-10  &$300$\\
				\hline
			\end{tabular}%
		}
	}
		\\
		\subfloat[
		\textbf{Proxy-free search.} Ablations for searching w/o proxy dataset.
		\label{tab:proxyfree-search} 
		]{
			\resizebox{1.0\linewidth}{!}{%
				\begin{tabular}{lccccc}
					\hline\\[-2.5ex]
					\multirow{2}{*}{Name}  & \multicolumn{2}{c} {Accuracy (\%)} & \multirow{2}{*}{\makecell{Search\\ Epoch}} & \multirow{2}{*}{\makecell{Search\\ Dataset}}  &\multirow{2}{*}{\makecell{Train\\ Dataset}}\\
					\cline{2-3}\\[-2.5ex]
					&Top-1  &Top-5  \\
					\hline\\[-2.5ex]
					transfer  &$69.6$  &$88.45$  &$600$  &CIFAR-10  &ImageNet-tiny\\
					\textbf{proxy-free}  &$\mathbf{70.53}$  &$88.54$  &$300$  &ImageNet-tiny  &ImageNet-tiny\\
					\hline
				\end{tabular}%
			}
		}
	\end{center}}\end{minipage}
}	
\end{table*}

\begin{figure}[!t]
	\centering
	\includegraphics[width=1.0\columnwidth]{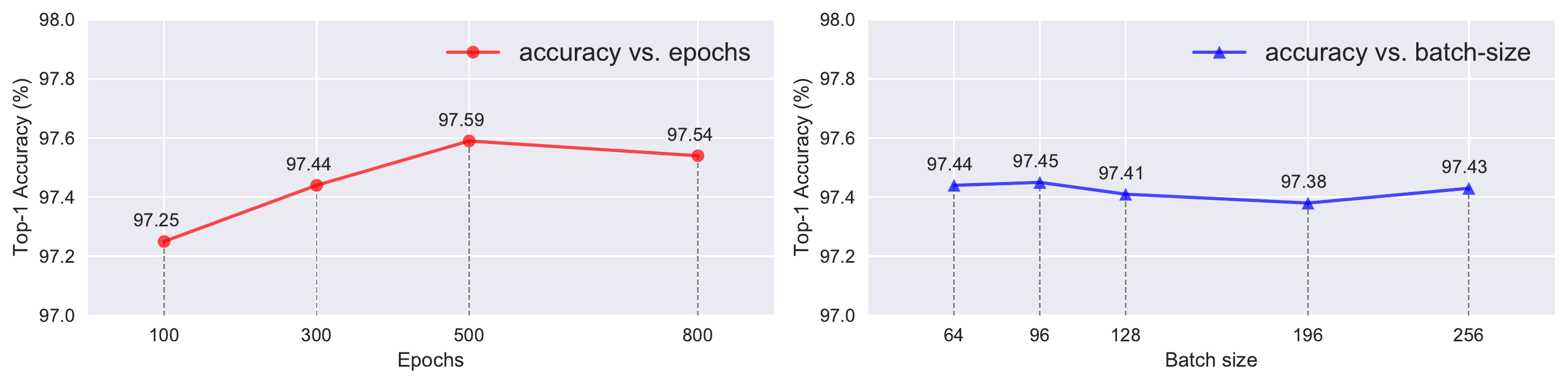}
	\caption{For ablations of searching epochs, we fix the batch-size to $96$ and increasing the searching epochs from $100$ to $800$. As for ablations of the batch-size, we fix the searching epoch as $300$ and scale batch-size from $64$ and $256$.}	
	\label{fig:epoch-bs}
\end{figure}

\noindent\textbf{Time span of FPP.} In this subsection, we study the role of time span in the FPP process and recommend the default setting. We allow a longer and a shorter FPP time span by adjusting the ratio to $[0.2, 0.2, 0.6]$ and $[0.2, 0.6, 0.2]$, respectively. Table~\ref{tab:fpp_ratio} compares the result of different time spans. It is shown that a balanced split between phase II and phase III strikes for a better result. Shorter FPP may result in the under-optimization of a single cell during the pruning phase. Longer FPP may squeeze the space of the normal bi-level optimization process, leading to potentially sub-optimal results. And a longer searching duration will relieve pressures of both phases by allowing more epochs at each phase and thus leads to a better result, as shown in Fig.~\ref{fig:epoch-bs}.

\noindent\textbf{Dropout for skip-connection.} Here we add a dropout rate for skip-connection during the lower-level optimization process. As shown in Table~\ref{tab:dropout}, using a $p=0.2$ (result from a coarse grid search) improves the overall quality of the architecture. However, the architecture quality is relatively sensitive to the dropout rate as the skip-connection is a crucial component to prevent gradient vanishing issue. Over-suppressing of the skip-connection can hamper the performance. 

\noindent\textbf{Proxy-free search.}\label{sec:proxy-free}
It is well-established~\cite{imagenet_pretrain_notwork,transfer_to_medical,SimCLR} that the performance of transfer learning drops when the gap between the target domain and the pre-trained domain is large. Given these priors, we focus on how proxy-free search can benefit the architecture quality. We use ImageNet-tiny~\cite{ImageNet} as the target domain and treat CIFAR-10~\cite{CIFAR} as the proxy domain. Then we carry out transfer learning and proxy-free search. We doubled the search epochs on CIFAR-10 to keep the same iterations used during the search (ImageNet-tiny contains $10^5$ images with size $64\times 64$, and CIFAR-10 contains $5\times 10^4$ images with size $32\times 32$ in the training dataset). So the only difference between the two settings is the domain gap (image dimension, context, etc.) itself. As shown in Table~\ref{tab:proxyfree-search}, proxy-free search results in a better architecture quality. This result suggests that given the existence of labeled proxy datasets, it is still favorable to search on the target domain directly without the label.

\noindent\textbf{Impact of epochs and batch-size.}\label{sec:epoch-bs} As verified in self-supervised weight pre-training~\cite{SimCLR,MOCO,BYOL}, network weights benefit from longer training and larger mini-batch size. Therefore, we examine how these two hyper-parameters affect the architecture searching process. As displayed in Fig.~\ref{fig:epoch-bs}, in consent with SSL weight pre-training, the quality of searched architecture also improves as the searching duration increases. It saturates around $500$ epochs on CIFAR-10/100. This result encourages a longer searching epoch and manifests that our design does not suffer from the mode collapse issue~\cite{darts_plus,ProxylessNAS}. And even searching for only $100$ epochs (cost $0.4$ GPU-day) on CIFAR-10, SSWP-NAS still surpasses DARTS ($1_{st}$ order)~\cite{DARTS}. Dissimilar to weight pre-training, the searched architecture does not take advantage of a larger mini-batch size in our approach referencing Fig.~\ref{fig:epoch-bs}. As a result, one can use SSWP-NAS reliably under the limited GPU memory without worrying about the degeneration of architecture quality. However, this conclusion is only effective under the scenario where the magnitude of mini-batch size is a few hundred. We do not verify the impact of mini-batch size in thousands as used in several self-supervised weight-pretraining works~\cite{SimCLR,MOCO,BYOL,simsiam}.  

\subsection{Weight-preserving Benefits Semi-supervised Learning}\label{sec:concomitant-weight} 

In this section, we use both the searched architecture and concomitant weights from SSWP-NAS and probe their performances under semi-supervised scenarios. In particular, we first search using SSWP-NAS for $500$ epochs with batch-size $96$ on CIFAR dataset (train data). Then we gradually reduce the labeled training data from $90\%$ to $10\%$ with step-size $20\%$ to mimic semi-supervised learning scenarios. Two baselines are included to demonstrate the effectiveness of concomitant weights. The first baseline noted as \textit{random initialization} only uses the architecture. It is then trained from scratch using the given data ratio with a random initialization. The second baseline,\textit{simclr pre-train}, in which we take the architecture searched by SSWP-NAS and pre-train it using SimCLR~\cite{SimCLR} for another $500$ epochs using the same batch-size. The second baseline corresponds to a common two-stage framework without weight-preserving property. All three methods share the same architecture and the only difference is how they are initialized. We run the above settings on both CIFAR-10 and CIFAR-100.

\begin{figure}[t!]
	\centering
	\includegraphics[width=1.0\columnwidth]{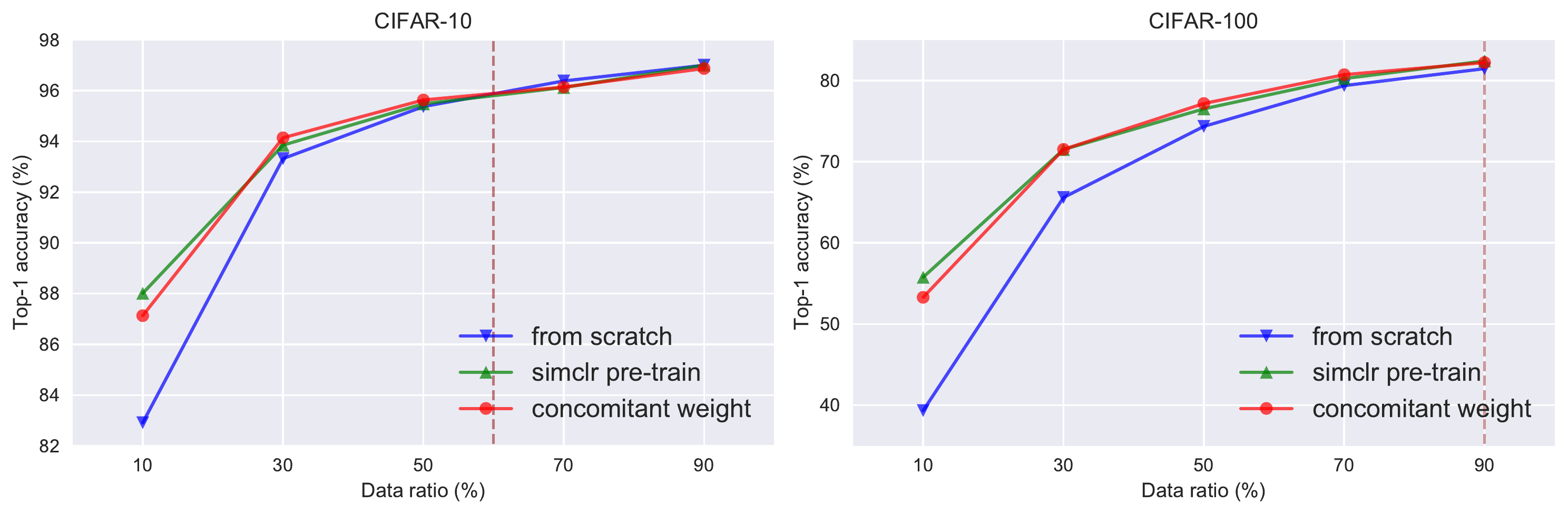}
	\caption{Performances of the random initialization, simclr weight pre-train and concomitant weights on CIFAR-10/100 under different training data proportions. }	
	\label{fig:weight-preserve}
\end{figure}

As shown in Fig.~\ref{fig:weight-preserve}, concomitant weights transfer clear positive information by outperforming the random initialization significantly. The gap between the concomitant weights and random initialization is bridged when using around $60\%$ train data on CIFAR-10. However, this benefit diminishes not until around $90\%$ of the train data on CIFAR-100. This result agrees with our intuition that self-supervised weight pre-training contributes more when labels are relatively scarce for each category, and the task is more challenging. More importantly, concomitant weights also surpass the two-stage framework pre-trained using SimCLR~\cite{SimCLR} except for the extremely scare data setting (with only $10\%$ data). On the one hand, this result substantiates our contribution by merging the common two-stage pipeline into one-stage; on the other hand, it further suggests an interesting potential that evolving both architecture and weights simultaneously may serve a better paradigm than the current isolated manner. 

\begin{table*}[t!]
	\centering{
	\caption{Ablation studies regarding concomitant weights.}
	\begin{minipage}[t]{0.47\textwidth}
	\begin{center}
		\subfloat[Comparisons of the two-stage pre-trained weights with the concomitant weights under different learning rates.
		\label{tab:weight-lr}]{
		\resizebox{1.0\textwidth}{!}{%
			\begin{tabular}{cccccc}
				\hline\\[-2.5ex]
				\multirow{3}{*}{\makecell{Learning \\rate}}  & \multicolumn{2}{c} {Accuracy (\%)} & \multirow{3}{*}{\makecell{Train data\\ ratio (\%)}} & \multirow{3}{*}{\makecell{Dataset}} \\
				\cline{2-3}\\[-2ex]
				&\makecell{SimCLR \\ Pre-train}  &\makecell{Concomitant \\ Weight}  \\
				\hline\\[-2.5ex]
				$0.01$  &$75.82$ &$73.18$&$50$\%  &CIFAR-100 \\
				$0.025$  &$\mathbf{76.1}$  &$74.96$  &$50$\%  &CIFAR-100  \\
				$0.1$  &$76.0$  &$77.2$  &$50$\%  &CIFAR-100  \\
				$0.2$  &$75.99$  &$\mathbf{77.54}$  &$50$\%  &CIFAR-100  \\
				$0.5$  &$74.18$  &$75.7$ &$50$\%  &CIFAR-100  \\
				\hline
			\end{tabular}%
		}
	}
	\end{center}
\end{minipage}
\hspace{1em}
\begin{minipage}[t]{0.47\textwidth}
		\begin{center}
		\subfloat[Impact of dropout and FPP on concomitant weights.
		\label{tab:weight-ablation}]{
		\resizebox{1.0\columnwidth}{!}{%
			\begin{tabular}{ccccc}
				\hline\\[-2.5ex]
				FPP  & Dropout & Relative gain & Train data ratio & Dataset \\
				\hline\\[-2ex]
				\XSolidBrush  &\XSolidBrush & - & 50\% & CIFAR-100 \\
				\XSolidBrush  &\Checkmark & $0.23$ & 50\% & CIFAR-100 \\
				\Checkmark   &\XSolidBrush & $0.01$ & 50\% & CIFAR-100 \\
				\Checkmark   &\Checkmark &$0.19$ & 50\% & CIFAR-100 \\
				\hline
			\end{tabular}%
		}
	}
	\end{center}
\end{minipage}
}
\end{table*}

To better understand the differences between two-stage pre-trained weights and our concomitant weights, we further verify their reactions to different learning rates. Here we use $50\%$ data from CIFAR-100 as a proxy-task. Table~\ref{tab:weight-lr} shows that, unlike the typical two-stage pre-trained weights, concomitant weights consistently enjoy a larger learning rate. This result implies different statistical distributions between the concomitant weights and two-stage pre-trained weights. 

Finally, we also investigate the impact of proposed modules on the quality of concomitant weights. Since architecture and concomitant weights are evolved simultaneously, we can not drive to the conclusion by direct comparing the accuracy. To this end, we use \textit{relative gain} to isolate the impact of the target operation towards the concomitant weights. $relative~ gain= (acc^{w}_{f} - acc^{w}_{s})- (acc^{o}_{f} - acc^{o}_{s})$ where superscript $w/o$ denotes with or without target operation. Subscript $f$ and $s$ represents fine-tuning and train from scratch, respectively. By doing so, we offset the influence of the architecture structure and focus on concomitant weights. As exhibited in Table~\ref{tab:weight-ablation}, dropout improves the quality of concomitant weights, this result agree with our assumption that insufficient update of parameterized operation may hinder the quality of the concomitant weights. And according to experiments, FPP does not have a clear impact on overall quality of concomitant weights.


\section{Conclusion} 
In this work, Instead of trying to further reduce the computational overhead of the search process, the proposed SSWP-NAS strikes for a simplified workflow of NAS. It enjoys both self-supervising and weight-preserving two properties. Experiments show that self-supervised learning consistently benefits SSWP-NAS, and the concomitant weights successfully merge the two-stage framework into the one stage. Comprehensive ablation studies substantiate the effectiveness of our proposed designs. For future work, it is important to probe how architecture and concomitant weights can boostrap each other. And it is also compelling to design the new self-supervised paradigm specifically for joint-optimization of architecture and concomitant weights. From the practical consideration, enabling multi-objective and hardware-aware learning would be useful. 

%
%
\bibliographystyle{splncs04}
\bibliography{egbib}

\section{Supplementary Materials}
\noindent We abbreviate self-supervised learning and supervised learning as SSL and SL for simplicity. 
	
	\subsection{Detailed Hyper-parameters}
	Detailed hyper-parameters used for both search and train stages are listed in Table~\ref{tab:hyper-prarmeter}. We follow the same suggestions as described in DARTS~\cite{DARTS} and ProxylessNAS~\cite{ProxylessNAS}.
	Variable $w$ refers to the lower-level parameter (operation weights) and $\alpha$ denotes upper-level parameter (architecture weights). To speed up the training process for ImageNet, we also employ the distributed data-parallel, the automatic mixed precision, and the synchronized batch normalization techniques implemented in pytorch~\cite{pytorch} framework.
	\begin{table*}[h]
		\caption{Detailed hyper-parameters used across the experiments.}
		\label{tab:hyper-prarmeter}
		\begin{center}
			\resizebox{1.0\linewidth}{!}{%
				\begin{tabular}{cccccc}
					\hline\\[-2.5ex]
					\multirow{2}{*}{\makecell{Hyper-\\parameters}}  & \multicolumn{2}{c} {CIFAR-10/100} & 
					\multicolumn{2}{c} {ImageNet}\\
					\cline{2-3}\cline{4-5}\\[-2.5ex]
					&Search  &Train  &Search  &Train \\
					\hline\\[-2.5ex]
					batch size & $96$ & $128$ & $256$ & $1024$  \\
					learning rate ($w$) & $0.025$ & $0.025$ & $0.025$ & $0.4$  \\
					minimum learning rate ($w$)  & $0$ & $0$ & $0$ & $0$  \\
					optimizer ($w$) & sgd & sgd & sgd & sgd  \\
					scheduler ($w$) & CosineAnnearling & CosineAnnearling & CosineAnnearling & CosineAnnearling  \\
					momentum ($w$) & $0.9$ & $0.9$ & $0.9$ & $0.9$  \\
					weight decay ($w$) & $4\times10^{-5}$ & $3\times10^{-4}$ & $4\times10^{-5}$ & $3\times10^{-4}$  \\
					learning rate ($\alpha$) & $0.001$ & - & $0.001$ & -  \\
					optimizer ($\alpha$) & adam & - & adam & -  \\
					adam $\beta_1$ & $0$ & - & $0$ & -  \\
					adam $\beta_2$  & $0.99$ & - & $0.99$ & -  \\
					auxiliary weight & - & $0.4$  & - & $0.4$  \\
					cutout~\cite{cutout} length & - & $16$  & - & -  \\
					drop-path rate & - & $0.3$ & - & - \\
					\hline
				\end{tabular}%
			}
		\end{center}
	\end{table*}
	
	\subsection{Self-supervised architecture search benefits self-supervised weight pre-training}
	In this section, we show that architecture searched by self-supervised learning objective also benefits the typical self-supervised weight pre-training method. We first search two architectures using supervised and self-supervised learning objectives, respectively. Then we pre-train the two searched architectures using the SimCLR~\cite{SimCLR} framework. Finally, we conduct the standard linear probe experiments~\cite{SimCLR,MOCO,BYOL,simsiam} on pre-trained two networks with the same setting. As shown in Table~\ref{tab:ssl_boost_ssl}, SSL searched architecture also benefits the downstream SSL based weight pre-training in a typical two-stage workflow (architecture search and weight pre-training are treated separately). This result suggests that the performance of the architecture and its corresponding weights are correlated under the SSL framework. It is recommended to use SSL-based NAS instead of SL-based NAS when considering using a self-supervised weight pre-training method. 
	
	\begin{table}[h]
		\caption{Linear probe results for architectures searched with SSL and SL.}
		\label{tab:ssl_boost_ssl}
		\begin{center}
			\resizebox{0.6\columnwidth}{!}{%
				\begin{tabular}{ccccccc}
					\hline\\[-2.5ex]
					Name  & Accuracy (\%) & \makecell{Search\\ epoch} & \makecell{Pre-train\\ epoch} & \makecell{Search\\ dataset}  &\makecell{Pre-train\\ dataset}\\
					\hline\\[-2.5ex]
					SL  &$64.4$  &$300$  &$150$ &CIFAR-10  &CIFAR-10\\
					SSL  &$65.0$  &$300$  &$150$ &CIFAR-10 &CIFAR-10\\
					\hline
				\end{tabular}%
			}
		\end{center}
	\end{table}
	
	\subsection{Architecture Structure Searched by SSWP-NAS}
\begin{figure}[b]
		\begin{center}
			\includegraphics[width=1.0\textwidth]{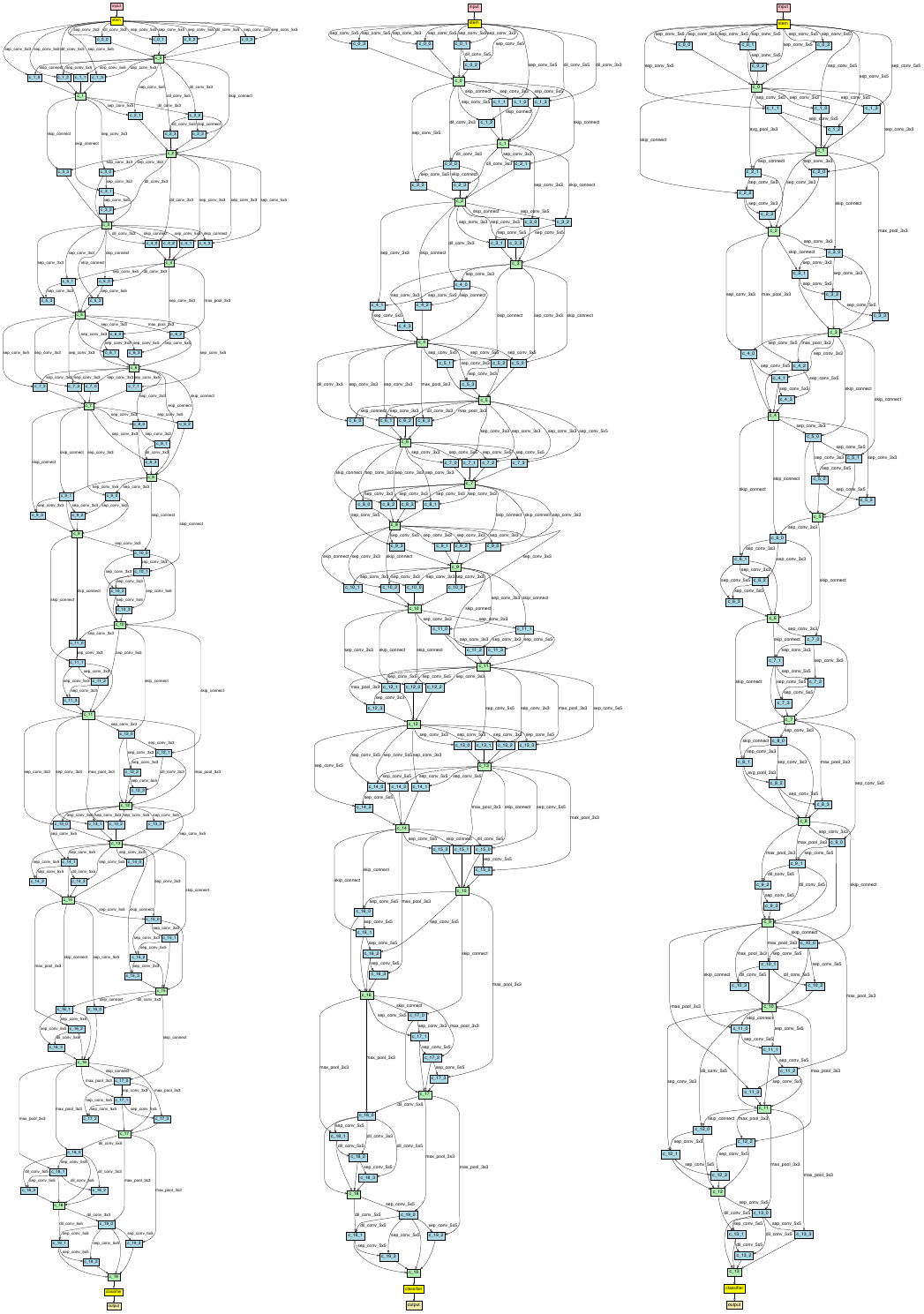}
		\end{center}
		\caption{Detailed architecture structures searched by SSWP-NAS. Architectures from left to right correspond to searching $300$ epochs on CIFAR-10/100, $500$ epochs on CIFAR-10/100, and $100$ epochs on ImageNet accordingly.}
		\label{fig:searched_architecture}
	\end{figure}

	Detailed architecture structures searched by SSWP-NAS are attached as Fig.~\ref{fig:searched_architecture}.
	
\end{document}